\documentclass[12pt]{article}
\usepackage{amsmath}
\usepackage{times}
\usepackage{graphicx}
\usepackage{color}
\usepackage{multirow}
\usepackage[authoryear]{natbib}
\usepackage{rotating}
\usepackage{bbm}
\usepackage{latexsym}
\newcommand{\captionfonts}{\normalsize}

\makeatletter
\long\def\@makecaption#1#2{%
  \vskip\abovecaptionskip
  \sbox\@tempboxa{{\captionfonts #1: #2}}%
  \ifdim \wd\@tempboxa >\hsize
    {\captionfonts #1: #2\par}
  \else
    \hbox to\hsize{\hfil\box\@tempboxa\hfil}%
  \fi
  \vskip\belowcaptionskip}
\makeatother

\begin{document}
\hspace{13.9cm}1

\ \vspace{20mm}\\

{\LARGE Emergence of Complex-Like Cells in a Temporal Product Network
  with Local Receptive Fields}

\ \\
{\bf \large Karol Gregor, Yann LeCun}\\
{Courant Institute of Mathematical Sciences, New York University}\\
%


\thispagestyle{empty}
\markboth{}{NC instructions}
\ \vspace{-0mm}\\
%
\begin{center} {\bf Abstract} \end{center}
We introduce a new neural architecture and an unsupervised algorithm
for learning invariant representations from temporal sequence of
images. The system uses two groups of complex cells whose outputs are
combined multiplicatively: one that represents the content of the
image, constrained to be constant over several consecutive frames, and
one that represents the precise location of features, which is allowed
to vary over time but constrained to be sparse.  The architecture uses
an encoder to extract features, and a decoder to reconstruct the input
from the features. The method was applied to patches extracted from
consecutive movie frames and produces orientation and frequency
selective units analogous to the complex cells in V1. An extension of
the method is proposed to train a network composed of units with local
receptive field spread over a large image of arbitrary size. A layer
of complex cells, subject to sparsity constraints, pool feature units
over overlapping local neighborhoods, which causes the feature units
to organize themselves into pinwheel patterns of orientation-selective
receptive fields, similar to those observed in the mammalian visual
cortex. A feed-forward encoder efficiently computes the feature
representation of full images.


\section{Introduction}

According to prevailing models, the mammalian visual cortex is
organized in a hierarchy of levels that encode increasingly higher
levels features from edges to object categories. The primary visual
area V1 contains {\em simple cells}, which primarily respond to
oriented edges at particular locations, orientations and frequencies,
and {\em complex cells} which appear to pool the outputs of multiple
simple cells over a range of locations.  Receptive fields similar to
simple cells have been shown to be produced by sparse coding
algorithms~\citep{olshausen1996emergence}.

The precise computation carried out by complex cells is not entirely
elucidated. One idea is that they pool simple cells that often respond
consecutively in time. In the so-called slow feature analysis
method~\citep{wiskott2002slow,berkes2005slow,bergstra22slow}, this is
achieved by penalizing the rate of change of unit activations as the
input varies. Another approach is to impose sparsity constraints on
complex cells that pool local groups of simple
cell~\citep{hyvarinen2001two,koray-cvpr-09}. This forces units within
a local group to learn similar filters that are often co-activated.
Another approach, applied to static image patches, is to model the
covariance of images, increasing the likelihood of features that
commonly occur together, forming their
representation~\citep{karklin2008emergence}. In \cite{NIPS2008_0200} the image is represented as a sparse model in terms amplitude and phase at the first layer and the second layer discovers translational invariants. Another structured model of video is \cite{berkes2009structured}.

This paper first demonstrates that sparse coding can be used to train
a network of locally-connected simple cells that operates on an image
of arbitrary size. The network is composed of a feed-forward encoder
which computes an approximation to the sparse features, and a decoder
which reconstructs the input from the sparse
features~\citep{koray-psd-08,koray-cvpr-09}. Unlike models such
convolutional networks~\citep{lecun-98}, the filters are not shared
across locations. The absence of shared weights is more biologically
plausible than convolutional models that assume identical replicas of
filters over the input field. The second section of the paper
introduces the use of sparsity criteria operating on local pools of
simple cells~\citep{hyvarinen2001two,koray-cvpr-09}. Since the
sparsity drives the number of active blocks to be small, simple cells
arrange themselves so that similar filter (which often fire together)
group themselves within pools. In a large, locally connected network,
this will result in orientation-selective, simple-cell filters that
are organized in pinwheel patterns, similar to those observed in the
primate's visual cortex. The third section introduces the temporal
product network that is designed to discover representations that are
invariant over multiple consecutive inputs. This produces units that
respond to edges of a given orientation and frequency but over a wide
range of positions, similar to the complex cells of V1. The model
includes a {\em feed-forward encoder architecture} that can produce
the internal representation through a simple feed-forward propagation,
without requiring an optimization process.

The architecture described below is applicable to any slowly-varying
sequential signal, but we will focus the discussion on temporal
sequences of images (video).

\section{Sparse Feature Learning in a Locally-Connected Network}

Predictive Sparse Decomposition
(PSD)~\citep{koray-psd-08,koray-cvpr-09} is based on Olshausen and
Field's sparse coding algorithm in which a decoding matrix is trained
so that image patches can be reconstructed linearly by multiplying the
decoding matrix by a sparse feature
vector~\citep{olshausen1996emergence}.  Unlike with sparse coding, PSD
contains an efficient {\em feed-forward encoder} (a non-linear
regressor), which is trained to map input images to approximate sparse
feature vectors from which the input can be linearly reconstructed.

As with sparse coding, PSD is normally trained on individual image
patches. Applying the resulting filters to a large image results in
redundant representations, because the learning algorithm contains no
mechanism to prevent a high degree of redundancy between outputs of
the same filter (or similar filters) at neighboring locations.  In
this section, we introduce a form of PSD that is applied to
locally-connected networks of units whose receptive fields are
uniformly spread over a large image.

\subsection{Sparse coding with an encoder.}

The basis of the PSD algorithm is Olshausen and Field's sparse coding
method for learning overcomplete basis
functions~\citep{olshausen1996emergence}. We denote by $X$ the input
vector (an image patch) of dimension $n_x$ and by $Z$ a (sparse)
feature vector of dimension $n_z$ from which the input is
reconstructed. The reconstructed input $\tilde{X}$ is produced through
a linear {\em decoder} $\tilde{X} = W^D \cdot Z$, where $W^D$ is an
$n_x \times n_z$ decoding matrix (or dictionary matrix) to be learned,
whose columns have norm 1 and are interpreted as basis vectors.
Given a decoding matrix $W^D$, sparse coding inference consists in
finding the feature vector $Z^*$ that minimizes the energy function
\begin{equation}
 E_{sc}(X,Z,W^D) = ||X-W^D \cdot Z||^2 + \alpha |Z|
\end{equation}
where $|Z|$ denotes the ${\rm L}_1$ norm of $Z$ (sum of absolute
values of the components). The positive constant $\alpha$ controls the
sparsity penalty.
\begin{equation}
 Z^* = {\rm argmin}_z E_{sc}(X,Z,W^D)
\end{equation}
The learning algorithm uses a gradient-based method to find the matrix
$W^D$ that minimizes the average of the following energy function over
a training set of input vectors $X$
\begin{equation}
 F_{sc}(X,W^D) = {\rm min}_Z E_{sc}(X,Z,W^D)
\end{equation}
In PSD, a parameterized encoder function ${\rm Enc(X,W)}$ is trained
to compute a prediction $\tilde{Z}$ of the optimal sparse vector
$Z^*$. In its simplest form, the encoder takes the form
\begin{equation}
 \tilde{Z} = {\rm Enc}(X,W) = D \tanh (W^E \cdot X + B)
\end{equation}
where $W$ collectively denotes the $n_z \times n_x$ encoding matrix
$W^E$, the $n_z \times n_z$ diagonal matrix $D$, and the $n_z$
dimensional bias vector $B$. In PSD, the encoder and decoder are
trained simultaneously. The optimal code $Z^*$ minimizes the following
energy function
\begin{equation}
 E_{psd}(X,Z,W^D,W) = ||X-W^D \cdot Z||^2 + ||Z-{\rm Enc}(X,W)||^2 + \alpha |Z|
\end{equation}
As with Sparse Coding, the PSD training procedure uses a
gradient-based method to find the $W^D$ and $W$ that minimize the
following objective function averaged over a training set of input
vectors
\begin{equation}
 F_{psd}(X,W^D,W) = {\rm min}_Z  E_{psd}(X,Z,W^D,W)
\end{equation}
An iteration of the training procedure is as follows. Given an input
vector $X$ and the current parameters, compute $\tilde{Z} = {\rm
  Enc}(X,W)$. Then initialize $Z=\tilde{Z}$, and find the $Z^*$ that
minimizes $E_{psd}(X,Z,W^D,W)$, using gradient descent or some other
iterative method.  Update the parameters $W^D$, $W^E$, $D$, and $B$ so
as to lower $F_{psd}(X,W^D,W) = E_{psd}(X,Z^*,W^D,W)$, using a step of
stochastic gradient descent. finally, re-normalize the columns of
$W^D$ to unit norm.  The normalization of $W^D$ is necessary to
prevent singular solutions in which $|Z|$ is very small, and $W^D$
very large. After training on natural image patches, the columns of
$W^D$ and the rows of $W^E$ become oriented edge detectors. The
inferred $Z^*$ for a typical image patch will be sparse, and the
predicted $\tilde{Z}$ will be quite close to the optimal $Z^*$ for any
$X$ near the manifold of high training samples density. The encoder
function provides a very efficient (feed-forward) way to produce an
approximation of the optimal $Z$. We interpret the rows of $W^E$ as
filters (or receptive fields) and the components of $\tilde{Z}$ as
simple cell activations.

\subsection{Locally-connected network.}

While the original PSD method is trained on individual patches, our
aim is to train an entire set of local filters over a large image
using PSD. We must point out that filters with different receptive
fields {\em are not constrained to be identical}.  This is very much
unlike ``convolutional'' approaches in which the weights of filters at
different locations are shared.  Basically, a given simple cell (a
given component of $Z^P$) is connected only to a local receptive field
in the input image. Similarly, the corresponding component in $Z$ is
connected to the same ``projection field'' in the input through the
decoder. In general the receptive fields can have arbitrary shapes,
not all inputs need to be connected, and different location can have
different densities of simple cells (e.g. a density that geometrically
decreases with excentricity, as in the primates' visual systems). In
the simplest case used here, the connectivity is uniform: the simple
cells form a two-dimensional regular grid over the image. Each one is
connected to a square receptive field directly below it. The density
of simple cells can be set to be higher than that of the pixels.  This
produces an over-complete representation in which several simple cells
have the same receptive field (but different weights). The density of
simple cells can also be lower than that of the pixels, corresponding
to an under-complete representation in which adjacent receptive fields
are stepped by more than 1 pixel. The densities can be identical,
producing a one-to-one representation.

Formally, if $s_x$, $s_y$ are the integer coordinates of a simple
cell, then its receptive field has coordinates $(m_x, m'_x) \times
(m_y, m'_y)$ where $m_i = (\max(\mbox{floor}(s_i/\rho_i - P_i/2),0)$,
$m'_i = (\min (\mbox{floor}(s_i/\rho_i + P_i/2)$, where $P_x \times
P_y$ is the size of the neighborhood, $\rho_i$ are the densities of
simple cells in the two directions, $m_i \in [0, N_i \rho]$,
$N_x\times N_y$ is the image size and $i=x,y$.

This network is considerably smaller than a fully connected network
with the same size input. The number of connections goes from $C.N^4$
to $C.N^2 P^2$ where the image is of size $N \times N$ and the local
neighborhood of size $P \times P$, and $C$ is the overcompleteness
factor. This makes training tractable for large images. Arguably,
constraining the receptive fields to be local hardly reduces the
capacity of the system, since sparse coding algorithm end up learning
highly localized filters, and zeroing out most of the weights.

\subsubsection{Periodic replication.}

While our locally-connected network can be trained on images of
arbitrary sizes, there is little advantage to training it on images
that are larger than a small multiple of the receptive field
size. This is because the activations of simple cells that are away
from each other are essentially independent of each other. Conversely,
the activations of nearby simple cells depend on each other through
the minimization under sparsity: neighboring units compete to explain
the input, implementing a kind of ``explaining away'' mechanism.
Hence we {\em replicate} a ``tile'' of weights in periodic fashion
over the image. Other way to say this is that, in a locally connected
network, we share (tie) those weights together that are multiple of an
integer distance away from each in each direction. This allows us to
train on smaller size inputs, such as $79 \times 79$ pixels and apply
it to an arbitrarily large image. This sharing takes advantage of the
fact that the statistics of the image is the same at different
points. If this periodicity is the same as the local neighborhood, the
number of weights becomes $P^4$ - the same as that of the
corresponding image patch. If the periodicity is 1, the system reduces
to a convolutional layer.

Formally, let $W_{s_x,s_y,p_x,p_y}$ be the weight matrix element
between simple cell at $(s_x,s_y)$ and pixel $(p_x,p_y)$. Then
$W_{s_x,s_y,p_x,p_y} = W_{s_x+h_x \rho_x,s_y+h_y \rho_y,p_x+h_x,p_y +
  h_x}$ where $h_x$ and $h_y$ are integers. Numbers are $\rho_i = k_i$
for overcomplete, $\rho_i = 1/k_i$ for undercomplete and $\rho_i = 1$
for complete system (in particular direction) where $k_i$ are
integers, $i=x,y$ and the $h_i \rho_i$ are also required to be
integers. Note that for $h_x=h_y=1$ the network reduces to
convolutional neural network with $\rho_x \rho_y$ number of feature
maps.

\subsubsection{Boundary Effects.}

Units at the periphery of the network receive less than $P\times P$
inputs, hence must be treated differently from regular units. If the
image size on which the system is trained were very large, the effect
of these units on the training process would be negligible. But it is
more efficient and convenient to train on images that are as small as
possible, generally around $3P \times 3P$. Hence to avoid a adverse
effects of the boundary units on learning, their weights are not
shared with other units. With this method, there is no visible
artifacts on the weights of the bulk units when training on images of
size $3P \times 3P$ or greater.

\subsection{Input Data and Preprocessing}

The method was tested with two datasets. In the first, $100 \times
100$ pixel windows were extracted from the Berkeley image
dataset. Consecutive frames were produced by shifting the window over
the original image by 1 or 2 pixels in any direction.  For the second
set of input, short sequences of consecutive frames were extracted
from the movie ``A Beautiful Mind''. Results are reported for the
first dataset, but the results obtained with the second one were very
similar.

Before extracting the windows and feeding then to the network, each
image is preprocessed by removing the local mean (using a high pass
filter) and normalizing the standard deviation (contrast
normalization) as follows. First each pixel is replaced by its own
value minus a gaussian weighted average of its neighbors. Then the
pixel is divided by the gaussian-weighted standard deviation of its
neighbors. The width of both gaussians was $11.3$ pixels.
In the contrast normalization there was a smooth cutoff that rescales
pixels with small standard deviation less.

\subsubsection{Training.}

The training proceeds the same way as in the patch-based version
of~\cite{koray-psd-08}. The optimal code is found using gradient
descent with a fixed number of steps. The code inference takes
$N^2/P^2$ more computation than with patch-level training. To minimize
the ``batching effect'' due to weight sharing, the weights are updated
on the basis of the gradients contributed by units with a single
common receptive field. After this update, the optimal code is
adjusted with a small number of gradient descent iterations, and the
process is repeated for the next receptive field. This procedure
accelerates the training by making it more ``stochastic'' than if the
weights were updated using the gradient contributions accumulated over
the entire image.

\begin{figure}[h]
\hfill
\begin{center}
\includegraphics[width=5.8in]{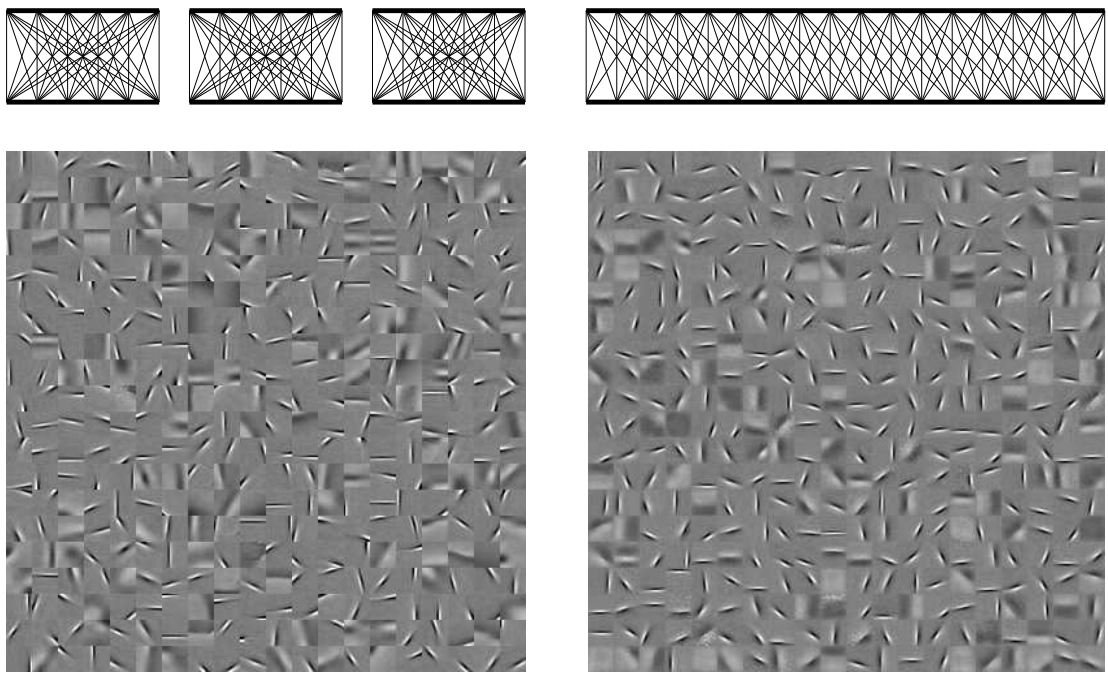}
\end{center}
\caption{Filters of a simple cell network (a) in
  patch training and (b) in periodic locally connected network. In
  both cases the network was complete and the patch/receptive field
  sizes were $20 \times 20$. In (b) the periodicity was $20 \times 20$
  and the system was trained on $79 \times 79$ images in the presence
  of boundary units. We see that in the latter case the entire supports of the
  filters are roughly within their receptive fields.}
\label{fig_patch_vs_local}
\end{figure}

\subsubsection{Results.}

Learned filters are shown in the Figure \ref{fig_patch_vs_local}. As
expected, oriented edge detectors are obtained, similar to those
obtained by training on patches. There is one significant difference:
in patch-based training, the filters have to cover all possible
locations of edges within the patch. By contrast, our system can
choose to use a unit with a neighboring receptive field to detect a
shifted edge, rather than covering all possible locations within a
single receptive field location. Hence, the units tend to cover the
space of location, orientations and frequencies in a considerably more
uniform fashion than if we simply replicate a system trained at the
patch level. Most of the filters are in fact centered within their
receptive field.

\subsubsection{A better encoder.}

The non-linearity used in the original PSD encoder is of the form $D
\tanh Y$ where $Y = WX+B$, and $D$ is a diagonal gain
matrix. Unfortunately, this encoder makes it very difficult for the
system to produce sparse output, since a zero output is in the
high-gain region of the $\tanh$ function. To produce sparse outputs,
the non-linearity would need a ``notch'' around zero, so that small
filter responses will be mapped to zero. Our solution is to use a
``double tanh'' function of the form $D (\tanh (Y+U) + \tanh (Y-U))$
where $U$ is a learned parameter that determines the width of the
``notch''. The prediction error is empirically better by about factor of two for complete network with this double-tanh than with the regular tanh.

\section{Pinwheel patterns through group sparsity.}

Hubel and Wiesel's classic work showed that oriented edge detectors in
V1 are organized in topographic maps such that neighboring cells
respond to similar orientations or frequencies at nearby
locations. Local groups of units can be pooled by complex cells with
responses that are invariant to small transformation of the
inputs. Hyvarinen and Hoyer have proposed to use group sparsity
constraints in a sparse reconstruction framework to force similar
filters to gather within groups~\citep{hyvarinen2001two}. The outputs
of units in a group are pooled by complex cells. Kavukcuoglu et
al. have proposed a modification of the PSD method that uses this idea
to produce invariant complex cells~\citep{koray-cvpr-09}. Here, we
propose to use the same idea to produce topographic maps over real
space: filters that are nearby in real space will also detect similar
features.  The new sparsity criterion for a single pool is:
\begin{equation}
E_{\mbox{sparsity}} = \alpha \sum_{r} \sqrt{ \sum_\delta Z_{r+\delta}^2 \exp (- \delta^2 /2 \sigma^2)}
\end{equation}
where $r=(x,y)$ is the vector of the coordinates of a simple cell and
$\delta$ is an integer vector. The overall criterion is the sum of
these over the entire domain (the pools overlap). This term tends to
minimize the number of pools that have active units, but does not
prevent multiple units from being simultaneously active within a pool.
Hence pools tend to regroup filters that tend to fire together.

\begin{figure}[h]
\hfill
\begin{center}
\includegraphics[width=4.5in]{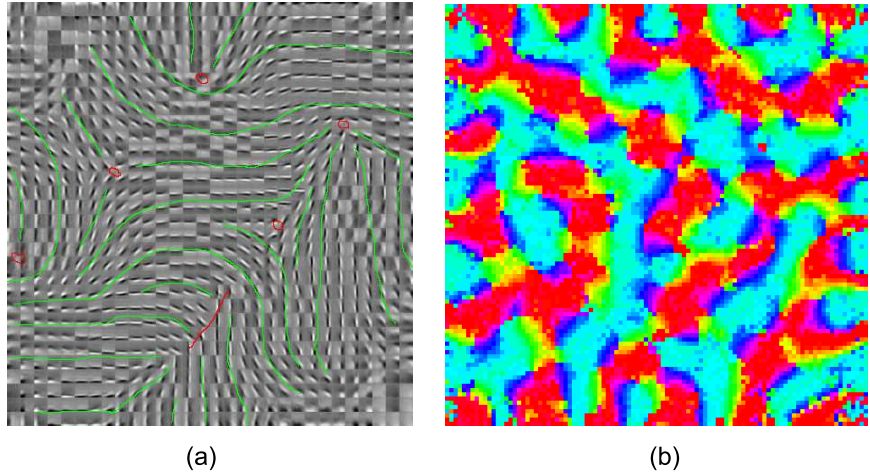}
\end{center}
\caption{(a) Filters of periodic locally connected network with
  pooling of simple cells over local neighborhoods (see text). The red
  circles (lines) denote locations of the topological defects. (b)
  Non-periodic locally connected network of size of size $100 \times
  100$ with pooling over local neighborhoods. The hue of each pixel
  indicates the orientation of the filter at each location. The
  orientations are estimated by fitting Gabor filters to each simple
  cell filter.}
\label{fig_pinwheel}
\end{figure}

We can apply this to a locally connected network in a natural way, as
the simple cells are already distributed on a two dimensional
grid. The result for a periodic locally connected network with local
neighborhood of size $15\times 15$, periodicity $20 \times 20$,
$4\times$ over-complete is shown in the Figure \ref{fig_pinwheel}a. We
see that the network puts the filters of the similar orientation and
frequency close to each other. Due to the topology of putting
orientations in the periodic grid, it is impossible to have smooth
transitions everywhere, which results in point topological defects -
and pinwheels patterns around them, familiar to
neurophysiologists. These are clearly visible in the Figure
\ref{fig_pinwheel}a which are marked by red circles/line.

In the periodic network, these have to fit periodically into the
square grid (on a torus). There is no periodicity in the brain, and
one has the usual maps over the whole area of
V1~\citep{obermayer1993geometry,crair1997ocular} with pinwheels distributed in somewhat
random non-periodic fashion (on a randomly deformed grid). This is
easily implemented here, in the locally connected network without
periodicity. We took the input of size $100 \times 100$ with local
neighborhoods of sizes $20 \times 20$ and a complete case ($\rho_i =
1$). At the location of each filter we draw a pixel with a color whose
hue is proportional to the orientation. This is shown in the Figure
\ref{fig_pinwheel}b. It very much resembles the maps obtained from the
monkey cortex in the reference \citep{obermayer1993geometry}.

\section{Temporal product network for Invariant Representations}

While the elementary feature detection performed by simple cells is a
good first step, perception tasks require higher-level features with
invariance properties. In the standard model of early vision, complex
cells pool similar features in a local neighborhood to produce locally
invariant features. The process eliminates some information.

The present section introduces an alternative method to learn complex
cells that are invariant to small transformations of the inputs. The
method preserves all the information in the image by separately
encoding the ``what'' and the ``where'' information.  As with slow
feature analysis~\citep{wiskott2002slow}, the main idea is
to exploit temporal constancy in video sequences. The system is again
built around the encoder-decoder concept, but the encoder and the
decoder are quite different from PSD's. The key ingredient is
multiplicative interactions. Hinton and his collaborators have
proposed several temporal models that use 3-way multiplicative
interactions~\citep{brown2001products,taylor2009factored}. Our model is
different in that two state vector of identical sizes are multiplied
term by term.

The product network described here operates on the output of simple
cells described in the previous sections, trained beforehand. More
precisely, the input to the product network are the absolute values of
the simple cell activations produced by the encoder discussed above.
The results are given with simple cells trained with ``simple''
sparsity, but the results obtained with the simple cells trained with
group sparsity are qualitatively similar.

\subsubsection{Separating the ``What'' from the ``Where''.}
The basic idea is to split the complex cells into two complementary
groups: invariant cells and location cells. The state of the invariant
cells is constrained to be constant for several consecutive frames
from a video. These cells will encode the content of the frames,
regardless of the position at which the content appears in the
frame. The complementary location cells encode the location of the
content, which may change between frames. The two codes cooperate to
reconstruct the input.

As an example, let us consider each input to be an edge at a
particular orientation that moves over time. Different simple cells
(at different locations) would respond to each frame. For simplicity
we can imagine that there is one simple cell active for each
edge/frame, though in reality the representation, while sparse, has
multiple active cells.  After training, each invariant complex cell
would respond to edges of a particular orientation at several
positions. During reconstruction, it would reconstruct all these at
every frame (the values of the simple cells corresponding to these
edges). Each complementary (position) cell would respond to edges at a
certain position but of various orientations. Different complementary
cells would be active at different time frames. At a given frame, an
active complementary cell would reconstruct the input edge at that
frame along with edges of other orientations at that position that it
is connected to. Taking the product of the reconstructions coming from
the invariant cells and the complementary cells gives the desired
input edge. Thus the input is translated into a more useful
representation: the orientation and the location.

\subsubsection{Encoder and decoder architectures.}

\begin{figure}[h]
\hfill
\begin{center}
\includegraphics[width=5in]{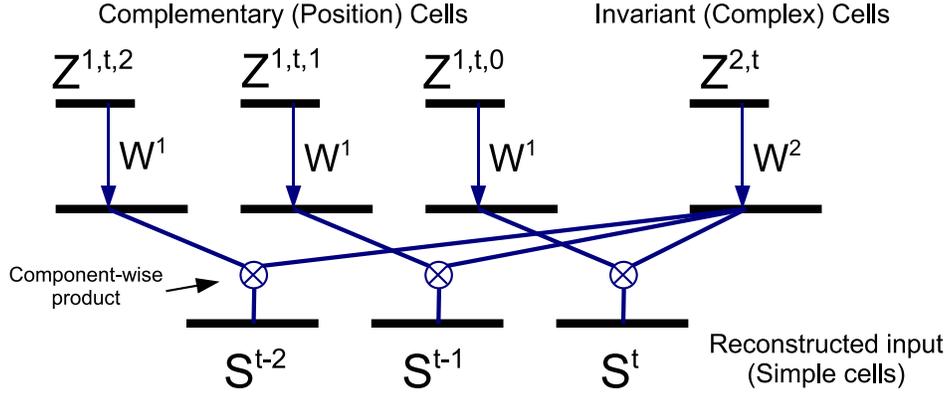}
\end{center}
\caption{Architecture of the decoder for the temporal product
  network. At a given time $t$, several frames are considered -
  the current one and several consecutive previous ones. The complex cell
  are divided into two groups: the invariant complex cells $Z^2$, and
  the position complex cells $Z^1$. The $Z^1$ cells form a sequence
  having different values at different frames. There is only one group
  of $Z^2$ cells, which is common to all frames (but varies at
  different times $t$ along with $Z^1$). The pairs of vectors $Z^1$
  and $Z^2$ (for each time delay) are multiplied by trainable matrices
  $W^1$ and $W^2$ and the results are multiplied term by term to
  produce the simple cell feature vector sequence $S$. These are then
  propagated through the simple cell linear decoder (basis vectors,
  not shown) to reconstruct the input image.}
\label{fig_tpn_dec}
\end{figure}

The detailed architecture of the decoder is given in figure
\ref{fig_tpn_dec}. Let the input to the temporal product network at
time $t$ be $S^t$ (the values of the simple cells). At this time $t$
we consider $N_\tau$ frames - the current one and the consecutive
previous ones. Let the invariant code be denoted by $Z^{2,t}$. At this
$t$ there is one complementary code for each time frame, denoted by
$Z^{1,t,\tau}$ where $\tau = 0,\ldots,N_\tau-1$. The invariant code
tries to turn on all the related simple cells at these time frames,
and the complementary code selects the correct one at each time
frame. The reconstructed input (the decoder operation) for the
$S^{t-\tau}$ at time $t$ is
\begin{equation}
S^{R,t,t-\tau} = \sqrt{(W^{D,1} \cdot Z^{1,t,\tau}) \times (W^{D,2} \cdot Z^{2,t})}.
\label{eq_SR}
\end{equation}
Here the $W^{D,1}, W^{D,2}$ are matrices, the dot denotes the matrix
multiplication and the cross the term by term multiplication of the
vectors. The columns of $W^{D,1}$, $W^{D,2}$ are normalized and the
$Z^1$, $Z^2$ are non-negative. The energy to be minimized is
\begin{equation}
E = \sum_{\tau} (S^{t-\tau}-S^{R,t,t-\tau})^2 + \alpha_1 \sum_{\tau} |Z^{1,t,\tau}| + \alpha_2 |Z^{2,t}|
\label{eq_Etpn}
\end{equation}
where we typically have $\alpha_1 = \alpha_2 = 0.02$.

The form of equations (\ref{eq_SR},\ref{eq_Etpn}) is not arbitrary. In this paragraph we give three intuitive arguments from which this form follows. The first two arguments are the same as for the simple cell network.
First, the normalization of the columns - sum of the {\em squares} equals to one - relative to the power of sparsity: {\em one} - is what causes sparse representations to have a lower energy. There could be different powers but the normalization power needs to be greater then the sparsity power. Second the sparsity power should be one for the following reason. Imagine the power was larger then one and we have two filters which are similar. Then given an input that perfectly matches the first filter, the other filter would also turn on because with power greater then one it is advantageous to distribute activity among both. Furthermore, this would pull the filters together. On the other hand if power was smaller then one, a given input would tend to commit to one of the units even though the other would also be a good explanation, though this might be acceptable.
Third, there should be square root for the following reason. Imagine there wasn't and that we have an input that can be well reconstructed. If we start with a small code, the gradient from the first term would be small (proportional to the code) but the gradient from the second term would be constant. Thus we would end up with a zero code even though there is a perfectly good code that reconstructs the input. What we need is that the size of the gradient is independent of the magnitude of the starting code (assume it is nonzero). Square root has this property. We have done some experiments without the square root and obtained invariant filters as well, but their diversity is not as good as of those obtained with the square root, especially in the locally connected network.

The encoder module is defined as follows:
\begin{eqnarray}
Z^{P,1,t,\tau} &=& D^1 (\tanh (W^{E,1} S^{t-\tau}+B^1+U^1) + \tanh (W^{E,1} S^{t-\tau}+B^1-U^1)) \\
Z^{P,2,t} &=& \sum_{\tau} D^2 (\tanh (W^{E,2} S^{t-\tau}+B^2+U^2) + \tanh (W^{E,2} S^{t-\tau}+B^2-U^2))
\end{eqnarray}
where $W^E$'s are the encoder matrices, $B^1$,$B^2$,$D^1$,$D^2$ are
vectors and $U^1$,$U^2$ are scalars.

\subsubsection{Comparison to slow feature analysis.}
In temporal product network, at every times step we are inferring a
code for several time frames. When we move by one time step, new set
of codes will be inferred. This is different from the slow feature
analysis. There a problem is that after arbitrary number of steps you
get artifacts from the previous times. Here code is inferred only on
the fixed set of frames. In a transition period between two invariant
features the network might not be able to reconstruct the input
properly, but as soon as we are well into the new feature, the
reconstruction doesn't have any artifacts. In fact, it might be
possible, but we haven't tested it, that even in the transitional
period, the reconstruction works well as follows. Invariant units for
both invariant features would be on, but if the complementary
connections don't have overlaps, the reconstruction is good.

\begin{figure}[h]
\hfill
\begin{center}
\includegraphics[width=3in]{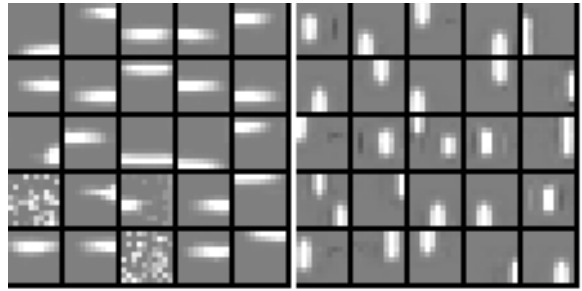}
\end{center}
\caption{Complex cell filters for a temporal product network trained
  on a toy problem. The input is a is a Gaussian bump moving to the
  right.  The filters of the invariant complex cells (left) and
  position complex cells (right) are shown. The input to the network
  is a $10\times10$ patch, with values given by a gaussian centered at
  $x,y$ of width $1.5$ pixels. For a given $y$ coordinate the gaussian
  moves to the right until it disappears from the image, at which
  point it is generated on the left at a random value of $y$. We see
  that the $Z^2$ cells are invariant to the $x$ position (direction of
  movement) in some range of $x$ and are therefore invariant to shifts
  that happen in time. The cells $Z^1$ are complementary and are
  invariant to the range of values $y$. In terms of direction of
  motion they extract position. More precisely their goal is to group
  inputs that are similar but are not along the direction of motion.}
\label{fig_moving_gauss}
\end{figure}

\section{Results.}
To help understand the function of the network, we show results for a
toy example in figure \ref{fig_moving_gauss}. The input to the
temporal product network is a $10\times10$ image patch whose values
are given by a gaussian of width $1.5$ pixels at a location $x,y$. The
gaussian moves to the right and when it disappears on the right, it
appears on the left at a randomly generated $y$. We see that the $Z^2$
cells are invariant to the $x$ position (direction of movement) in
some range of $x$ and are therefore invariant to shifts that happen in
time. The cells $Z^1$ are complementary and are invariant to the range
of values $y$. In terms of direction of motion they extract
position. More precisely their goal is to group inputs that are
similar but are not along the direction of motion.

\begin{figure}[h]
\hfill
\begin{center}
\includegraphics[width=5in]{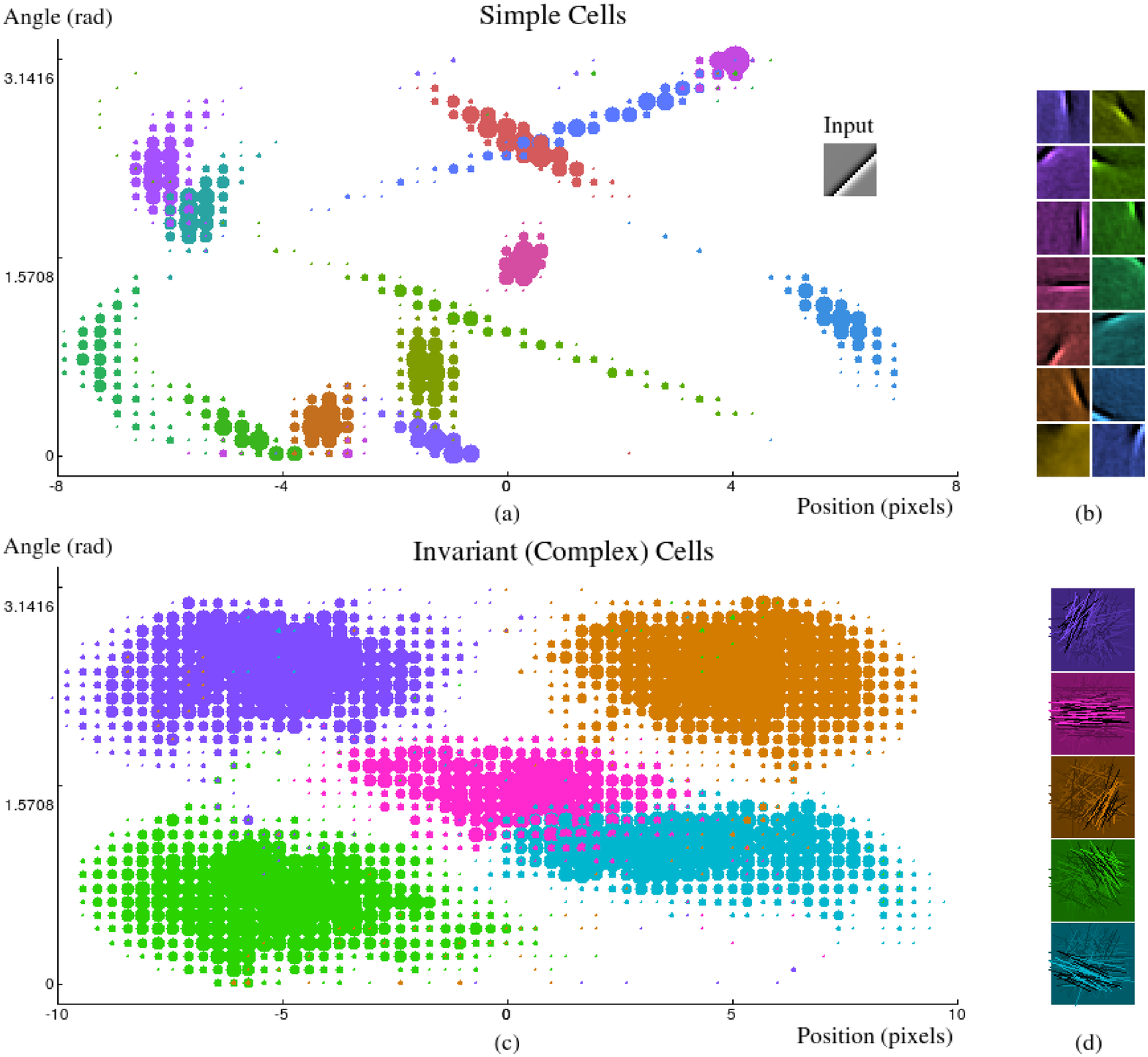}
\end{center}
\caption{(a) Responses of the simple cell filters shown in (b) to an
  edge, as a function of edge position (distance from the center) and
  orientation. Each color represents one simple cell, and the size of
  the bubble is proportional to the cell's activity. (c) Same but for
  the invariant cells shown in (d). (d) was obtained by fitting gabor
  function to simple cell filters, and plotting lines with magnitudes
  proportional to the connection between the invariant and simple
  cells and orientations and position obtained from the
  fit.}
\label{fig_activ}
\end{figure}

Now we discuss the realistic example of image training. We start with
the discussion of image patch training. The results for locally
connected network are below.
The Figure \ref{fig_activ}b shows a selection of the simple cell
filters. We fitted each of the simple cell filters with a Gabor
function, giving us among other parameters, orientation, position and
frequency of the filter. The Figure \ref{fig_activ}d shows a selection
of complex cells. Each line correspond to a simple cell filter, with
orientation and position obtained from the fit, and the intensity
proportional to the strength of the connection between the complex and
simple cells. We see that each invariant cell has strong connections
to edge detectors of similar orientation at a range of positions.

Next, we look at the responses of the simple and invariant cells to
moving edges. We parameterize an edge by orientation and position, the
later being the distance from the center of the input patch. The
responses of the simple cells are in the Figure \ref{fig_activ}a. Each
color correspond to a different simple cell and the size of the bubble
is proportional to the activity. Analogous graph for the invariant
cells is shown in the Figure \ref{fig_activ}c. We see that the
invariant cells respond to much larger range of positions then the
simple cells but a similar range of orientations. We also see that the
responses are quite smooth. (The edge was moving very slowly,
practically stationary. For faster moving edge the responses are even
smoother.)

\begin{figure}[h]
\hfill
\begin{center}
\includegraphics[width=4.5in]{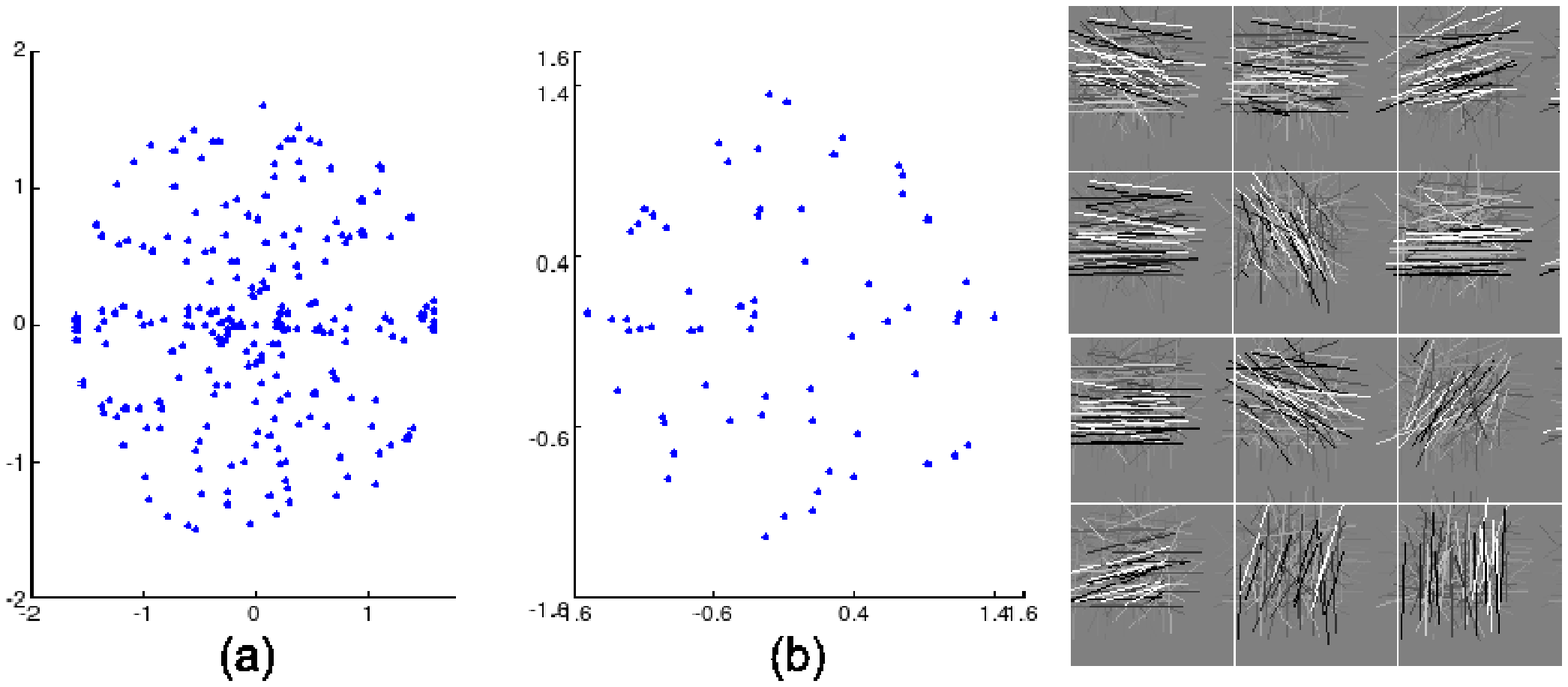}
\end{center}
\caption{(a) Frequency orientation plot for the gabor fits of the
  simple cell filters. The radius is proportional to the frequency and
  angle to twice the angle of orientation of the filter. (b) Same but
  for the invariant cells. The parameters were obtained as the
  weighted average of simple cell parameters with weights proportional
  to the square of the connections to simple cells. Right panel: A
  selection of line plots for invariant cell filters analogous to
  Figure \ref{fig_activ}d.}
\label{fig_SC_params}
\end{figure}

Next we discuss the diversity of the filters. For the system to
perform well it should have filters distributed evenly among
orientations and have a diversity of frequencies. This time we show
the results for the locally connected network. The simple cell network
used here has a local neighborhood of size $16 \times 16$, is
$4\times$ over-complete and with periodicity in the $x$ space of $8$
in each direction. The complex cell network has local neighborhood of
size $16 \times 16$, $4\times$ under-complete with respect to the
simple cell layer, with periodicity $16$ in each direction. The
orientation/frequency plot for the simple cells is in the Figure
\ref{fig_SC_params}a and for the complex cells in the Figure
\ref{fig_SC_params}b. The radius is the frequency and the orientation
is twice the angle from the $x$ axis. For the complex cell there is no
such number, but since they have strong connections to edges of
similar orientation and frequency we calculate the average weighted by
the square of this connection. We see that in both cases we have a
smooth distribution in the frequency/orientation space. This is where
the form (\ref{eq_SR}) of the reconstruction is important. Without the
square root for example the parameters are not so well distributed.

The final system that we obtain can be applied to large images and
used for fast image recognition as it contains a feed-forward pass
through the whole system. Let us recapitulate the computations
involved in the forward pass calculation of activities of complex
cells. The preprocessing contains two convolutions - for mean and
standard deviation removal. The simple cell calculation is not a
convolution but application of different filter at different point
(followed by nonlinearity). However, the computational cost is
equivalent to computing a number of convolutions equal to the
over-completeness of the system. Results are presented for a complete
(but not over-complete) system, hence the cost is equal to a single
convolution. The next level, which contains the complex cells, also
involves the application of different filters at each point, followed
by a nonlinearity. In this case we use four times under-complete
system and hence the computational cost is equivalent to one quarter
of a convolution. Afterwards we train logistic regression classifier.

\section{Efficiency of locally connected organization}

Convolutional net is a special case of a periodic locally connected net when the periodicity in the input space is one in each direction. This arrangement appears sensible since the input statistics is translationaly invariant. However if we have a limited computational capacity, there is only a limited number of filters in the image that we can use. It is likely that for a given filter (especially a low frequency one) it is enough to apply it with spacing of two or more pixels. This frees up additional resources to use for filters of different type and allows richer model. However the question is: What is the right spacing to use? It is likely that large spacing is enough for low frequency filters so the spacing is ideally filter dependent. However filters are elongated along one direction, so the ideal spacing is likely to be orientation dependent as well. Ideally the network should discover the correct allocation of filters on its own. This is precisely what locally connected network does. However with the locally connected network, we have lost translational invariance, for example we don't know which filters to pool together in the pooling layer of convolutional net. Hence we need to learn this pooling. This is what the temporal product network does. Further it can learn to pool appropriate slightly different orientations together since they are all activated for a given edge.

To test these ideas we trained locally connected network unsupervised on Berkeley images as described above. We used $20\times 20$ filters and complete simple cell layer (density of simple cells equals density of inputs). The number of computations of the feedforward pass is the same as that of one convolution. For the complex cell layer we used $20 \times 20$ filters and four times undercomplete system. The number of computations this time is that of a quarter of convolutions. Note that preprocessing contained two convolutions. We testing the performance on Caltech 101 dataset~\citep{fei2007learning} with 30 training images per category. The resulting performance was $51\%$. Performing a local subtraction and contrast normalization on the top layer improves it to $54\%$.

These results are not state of the art which is currently for systems of this class is $75\%$ \cite{boureau-cvpr-10}
(the system extracts sift features, then learns sparse dictionaries, pools and makes histograms). There are several system of this class \citep{lazebnik2006beyond,pinto2008real,pinto2009high,serre2007feedforward,yang-cvpr-09,lee2009convolutional,NIPS2009_0719}
but specifically convolutional nets achieve $67\%$ \citep{jarrett-iccv-09}. However this system should be compared to a single layer convolutional net, since it essentially consists of filters with nonlinearity (simple cells) and pooling (complex cells). Single layer neural network achieves about $54\%$ performance which is the same as this network. However convolutional net is much larger, it typically involves $64$ $9x9$ convolutions. Thus the locally connected net can achieve the same performance at lower computational cost. This gives the merit to the idea descried above that locally connected organization is more efficient then convolutional one. However more experimental evaluation is needed.

\section{Conclusion}

We have presented a new neural architecture that follows more closely
the kind of calculations performed by the visual cortex but which at
the same time can be used for real time object recognition. It is a
layered architecture. It's first layer is a locally connected version
of PSD architecture. It's main feature is that the weights are not
shared for nearby filters (but can be for filters at larger distances
for efficiency) and the geometry is smooth, e.i. contains no cuts. The
next layer features a new algorithm for invariance extraction from
temporal data. It's aim is to translate the input into two types of
information - the ``what'' information that is invariant and the
``where'' information that complements the invariant one. This layer
is also designed in a locally connected way. Both layers include
encoder that predicts the values of the cells in a fast feed-forward
fashion. Therefore by including one of the standard classifiers
(logistic regression in out case) the whole system can be used for
fast visual recognition. As the system is smaller then other ones
typically used, the recognition is faster but the performance is
lower. It is left for the future work to see how this performance can
be improved.

This architecture suggests that locally connected organization without sharing of nearby weights is more efficient the convolutional one because it allocates correct filters at every location, rather then applying the same filter unnecessarily often.

In the future we need to increase the performance of the system, use
more over-complete representations, train more layers of the system
and show conclusively, if true, that locally connected training is more efficient the convolutional one.

\subsection*{Acknowledgments}
We thank Koray Kavukcuoglu for useful discussions.

\clearpage
\bibliographystyle{apalike}
\bibliography{nc-tpn}

\end{document}